\pdfoutput=1

\documentclass[11pt]{article}

\usepackage[final]{acl}
\usepackage{times}
\usepackage{latexsym}
\usepackage[T1]{fontenc}
\usepackage[utf8]{inputenc}
\usepackage{microtype}
\usepackage{inconsolata}
\usepackage{graphicx}
\usepackage{xcolor,colortbl}
\usepackage{multicol}
\usepackage{multirow}
\usepackage{caption}
\usepackage{subcaption}
\usepackage{booktabs}
\usepackage{amssymb}
\usepackage{mathtools}
\usepackage{pifont}
\usepackage{xurl}
\usepackage{wrapfig}
\usepackage{enumitem}
\usepackage{float}
\usepackage{capt-of}

\usepackage{stackengine}
\usepackage{multirow}
\usepackage{array}

\definecolor{mygray}{gray}{0.9}

\usepackage[flushmargin]{footmisc}

\usepackage[linesnumbered,ruled,vlined]{algorithm2e}
\usepackage{algpseudocode}  
\usepackage{amsmath}  

\DeclareMathOperator*{\argmax}{arg\,max}
\newcommand{\eg}{\emph{e.g.}}

\newcommand{\ie}{\emph{i.e.}}

\definecolor{lightgray}{RGB}{233, 233, 233}

\title{Meta-Semantics Augmented Few-Shot Relational Learning}

\author{
Han Wu\textsuperscript{1}$^,$\textsuperscript{2} \and 
Jie Yin\textsuperscript{1}\thanks{Corresponding author} \\
\textsuperscript{1}The University of Sydney, Australia\\
\textsuperscript{2}Peking University, China\\
\texttt{\{han.wu, jie.yin@sydney.edu.au\}}}

\begin{document}
\maketitle
\begin{abstract}
Few-shot relational learning on knowledge graph (KGs) aims to perform reasoning over relations with only a few training examples. While current methods have focused primarily on leveraging specific relational information, rich semantics inherent in KGs have been largely overlooked. To bridge this gap, we propose PromptMeta, a novel prompted meta-learning framework that seamlessly integrates meta-semantics with relational information for few-shot relational learning. PromptMeta introduces two core innovations: (1) a Meta-Semantic Prompt (MSP) pool that learns and consolidates high-level meta-semantics shared across tasks, enabling effective knowledge transfer and adaptation to newly emerging relations; and (2) a learnable fusion mechanism that dynamically combines meta-semantics with task-specific relational information tailored to different few-shot tasks. Both components are optimized jointly with model parameters within a meta-learning framework. Extensive experiments and analyses on two real-world KG benchmarks validate the effectiveness of PromptMeta in adapting to new relations with limited supervision. 
\end{abstract}

\section{Introduction}

Knowledge Graphs (KGs) represent real-world knowledge as a collection of factual triplets. Each triplet in the form of (\texttt{head entity}, \texttt{relation}, \texttt{tail entity}) indicates a relationship between a head entity and a tail entity. Large-scale KGs such as Wikidata~\citep{wikidata}, NELL~\citep{nell}, YAGO~\citep{yago}, and Freebase~\citep{freebase} form the backbone of a myriad of AI-driven applications, including question answering~\citep{KG4QA}, Web search~\citep{KG4websearch}, and recommender systems~\citep{KG4RS}. Despite their strong utility, KGs are often highly incomplete due to their semi-automatic construction from unstructured sources and the continual emergence of new knowledge. This challenge is compounded by the long-tail distribution of data, where most relations are associated with only a handful of triplets. As a consequence, predicting missing triplets for these sparse, long-tail relations becomes particularly difficult, hindering KG-based inference and reasoning capabilities. 


To circumvent these challenges, few-shot relational learning (FSRL) has been proposed. Given an unseen relation and only a few support triplets, the goal of FSRL is to predict the missing tail entity for query triplets. The core challenge of FSRL lies in effectively learning meta-knowledge from the limited support triplets that can be generalized to the query set. Most state-of-the-art FSRL methods adopt a meta-learning framework~\cite{MetaR, MetaP, GANA, HiRe},  where a base model is pre-trained on a series of meta-training tasks and then fine-tuned using only a few support triplets to make predictions for new and unseen relations. 



Although meta-learning based methods for FRSL have achieved competitive results, they predominantly rely on KG embedding models like TransE~\citep{TransE} or DistMult~\citep{DistMult} to exploit relational information in KGs. These methods embed entities and relations into a latent space and then optimize a scoring function based on relational structures 
to learn a few-shot model. However, this structure-centric approach overlooks the rich semantic information inherent in KGs~\cite{MetaR,Cornell22}. For example, the relations \texttt{FatherOf} and \texttt{ParentOf}, are semantically linked as they both describe familial relationships. If \texttt{FatherOf} holds between two entities, one can infer that \texttt{ParentOf} is also valid, even with limited training examples. 
In few-shot settings, where only a few training triplets are available for each relation, leveraging such semantic relatedness is crucial for better generalizing by allowing the model to infer patterns 
beyond direct structural connections. However, how to effectively integrate semantic information into few-shot relational learning to improve model generalization remains largely unexplored.

To fill this gap, we propose \textbf{PromptMeta}, a novel prompted meta-learning framework for few-shot relational learning. PromptMeta integrates meta-semantic knowledge with relational information inherent in KGs to improve generalization to rare or new relations. Inspired by the recent success of prompting techniques in NLP~\cite{zhou-etal-2022-dual,ENG}, PromptMeta incorporates prompt learning into meta-representation learning for few-shot tasks, offering a distinctive approach to improving generalization. Its novelty lies in two aspects: First, we propose a Meta-Semantic Prompt (MSP) pool that dynamically learns and consolidates high-level meta-semantics shared across tasks, facilitating effective knowledge transfer. Second, we introduce a learnable fusion mechanism that effectively integrates these meta-semantics with task-specific relational information, tailored to individual tasks. Both components are jointly optimized with model parameters within a meta-learning framework. 
Our main contributions are as follows:
\begin{itemize}[left=0pt]
\vspace{-2pt}
\setlength\itemsep{-5pt}
    \item We propose \textbf{PromptMeta}, a novel prompted meta-learning framework that synergistically combines meta-semantics with relational information for few-shot relational learning, addressing a critical research gap in the field.
    \item Extensive experiments and analyses on real-world KG datasets demonstrate that PromptMeta significantly outperforms state-of-the-art methods, validating its superior adaptation capability.
    \item We construct and release pre-trained semantic entity embeddings on two widely used KG datasets, creating valuable resources for advancing future research on few-shot relational learning.
\end{itemize}

\section{Related Works}
\paragraph{KG Embedding Learning.}
KG relational learning, or KG completion, primarily relies on embedding methods that encode KG relational structures to embed entities and relations into a fixed low-dimensional latent space. These methods typically define a scoring function to measure the plausibility of triplets. Translation-based methods like TransE~\citep{TransE}, TransH~\citep{TransH}, and TransR~\citep{TransR} assume a translational relationship between entities and relations for learning a unified embedding space. ComplEx~\citep{ComplEx} and DistMult~\citep{DistMult} improve the modeling of relational patterns in a vector or complex space. Other works such as KG-BERT~\cite{KG-bert} and LASS~\cite{shen2022LASS} reframe KG completion as a sentence classification task and incorporate semantic information by fine-tuning pre-trained language models at high computational costs. KGE-SymCL~\cite{liang2023knowledge} introduces a KG contrastive learning framework that leverages symmetrical structure information to enhance the discriminative power of KG embedding models.
Nevertheless, these methods heavily rely on a large amount of training examples. Their performance deteriorates under few-shot settings, where only a handful of training triplets are available for each relation. 

\paragraph{Few-Shot Relational Learning.}
Existing few-shot relational learning methods fall into metric-learning based and meta-learning based approaches. Metric-learning based methods learn a metric that measures the similarities between the support and query triplets. GMatching~\citep{Gmatching} introduces one-shot settings, where a neighbor encoder with equal weights is used to generate entity embeddings and a matching network is used to compare similarity between support and query entity pairs. FSRL~\citep{FSRL}~generalizes to few-shot settings and adopts a recurrent autoencoder to aggregate few-shot instances in the support set, yet imposing an unrealistic sequential dependencies among support triplets.
FAAN~\citep{FAAN} employs a dynamic attention mechanism to improve one-hop neighbor aggregation. CSR~\cite{CSR22neurips} and SAFER~\cite{SAFER2024liu} extract subgraphs from support triplets and adapt them to the query set for prediction. 

Most current FSRL methods fall into the regime of model-agnostic meta-learning (MAML)~\cite{MAML}, which focuses on quickly adapting to new few-shot relations by pre-training a base model on prior tasks to obtain a better initialization for fine-tuning on unseen relations. MetaR~\cite{MetaR} learns relation-specific meta-information simply by averaging support triplet representations. MetaP~\cite{MetaP} employs a meta-pattern learning framework for one-shot relational learning. 
GANA~\cite{GANA} integrates meta-learning with TransH~\cite{TransH} and devises a gated, attentive neighbor aggregator to handle sparse neighbors. HiRe~\cite{HiRe} further exploits multi-granular relational information to learn meta-knowledge that better generalizes to unseen relations. RelAdapter~\cite{RelAdapter} enriches entity embeddings with pre-trained context information and applies a feedforward network to adapt relation-meta to target relations. 

The aforementioned FSRL methods primarily rely on relational information, but neglect rich semantics inherent in KGs. Our work addresses this important gap by effectively learning meta-semantic knowledge shared across tasks and integrating it with relational information for FSRL. In parallel, several studies explore few-shot learning and meta-learning with graph neural networks (GNNs)~\cite{yang2022few, mandal2022metalearning}. While related, they are not directly applicable to our setting, which requires modeling complex relational patterns in KGs for generalizable FSRL.



\paragraph{Prompt Learning on Graphs.}
Built upon the recent success in language model fine-tuning, prompting techniques are increasingly explored for tasks that combine large language models (LLMs) with structured knowledge sources such as graphs~\cite{cocolm}. Hard text prompts, commonly used in NLP tasks~\cite{zhou-etal-2022-dual,ENG}, prepend handcrafted instructions to an input text, guiding models to extract relevant task-relevant knowledge using masked language modeling for downstream tasks. For example, G-Prompt~\citep{huang2023promptbased} uses hard text prompts to extract task-relevant node features, which are then fed into a GNN layer for few-shot node classification. TAPE~\cite{he2024harnessing} prompts an LLM to retrieve textual explanations as features to enrich node representations for downstream GNN training. 
Alternatively, learnable prompts have been proposed to reduce the high engineering costs in prompt design. A common approach is to define a graph template shared by both pre-training and downstream tasks for better adaptation, often coupled with specific GNN layers. GraphPrompt~\cite{GraphPrompt} and HGPrompt~\cite{HGPrompt} are exemplars that integrate prompts as learnable vectors within a hidden GNN layer by modifying the \texttt{Readout} function or its input. 

Our prompt design fundamentally differs from existing text-based and GNN-based prompting via two key innovations: (1) a Meta-Semantic Prompt (MSP) pool to capture meta-semantics and (2) a learnable fusion token to integrate meta-semantics with relational information. These innovations enhance knowledge transfer and model adaptation, addressing the unique challenges of FSRL on KGs.



\section{Problem Statement}
A knowledge graph (KG) can be represented as a collection of triplets $\mathcal{G} = \{(h, r, t)\}$, where \(h, t \in \mathcal{E}\) denote the head and tail entities, and \(r \in \mathcal{R}\) denotes a relation. The task of few-shot relational learning (FSRL) aims to predict new triplets for a relation \(r\) given only a small set of support triplets. 

The FSRL is formulated as a task-based meta-learning problem. Each task \(\mathcal{T}_{r_i} = (\mathcal{S}_{r_i}, \mathcal{Q}_{r_i})\) corresponds to a specific relation \(r_i\). The support set, $\mathcal{S}_{r_i} = \{(h_k, r_i, t_k)\}_{k=1}^K$, consists of $K$ training triplets (i.e., $K=1$, 3 or 5), where \(h_k, t_k \in \mathcal{E}\). 
Given the query set \(\mathcal{Q}_{r_i} = \{(h_l', r_i, ?)\}\), where \(h_l' \in \mathcal{E}\), the goal is to predict the missing tail entity $?$. For each query triplet \((h_l', r_i, ?)\), a candidate set \(\mathcal{C}_{h_l', r_i}\) is provided, and the model aims to rank the true tail entity highest among all candidates.

During meta-training, the model learns from a set of tasks, each corresponding to a relation:
\begin{align}
\mathcal{D}_{\text{train}} = \{\mathcal{T}_{r_i} \mid r_i \in \mathcal{R}_{\text{train}}\}.\nonumber
\end{align}
The model is optimized to learn generalizable knowledge that enables fast adaptation to new tasks. During meta-testing, the model is evaluated on tasks associated with novel relations, denoted as:
\[
\mathcal{D}_{\text{test}} = \{\mathcal{T}_{r_j} \mid r_j \in \mathcal{R}_{\text{test}}\}.
\]
Notably, the relation sets involved in meta-training and meta-testing are disjoint:
$\mathcal{R}_{\text{train}} \cap \mathcal{R}_{\text{test}} = \varnothing$,
where $\mathcal{R}_{\text{train}} = \{ r_i\}_{i=1}^I$ and 
$\mathcal{R}_{\text{test}} = \{ r_j \}_{j = I+1}^J$. The goal of FSRL is to generalize to novel relations by leveraging knowledge learned during meta-training. The model is trained on the meta-training set \(\mathcal{D}_{\text{train}}\) to acquire generalizable knowledge, and then evaluated on the meta-testing set \(\mathcal{D}_{\text{test}}\) to assess its ability to adapt to unseen relations using only a few support triplets.

\section{The Proposed Method}
As discussed, the core challenges in FSRL are twofold: (1) enabling model adaptation from the support set---where task-specific information is provided but highly limited---to the query set; (2) learning transferable knowledge during meta-training to generalize to unseen relations in meta-testing. To tackle these challenges, our proposed \textbf{PromptMeta} integrates relational and semantic information to improve adaptation and generalization. It comprises three components: (1) context-aware neighbor aggregation, (2) a learnable meta-semantic prompt pool for knowledge transfer, and (3) relational and meta-semantic information fusion.
\begin{figure*}[ht]
    \centering   \includegraphics[width=0.88\linewidth]{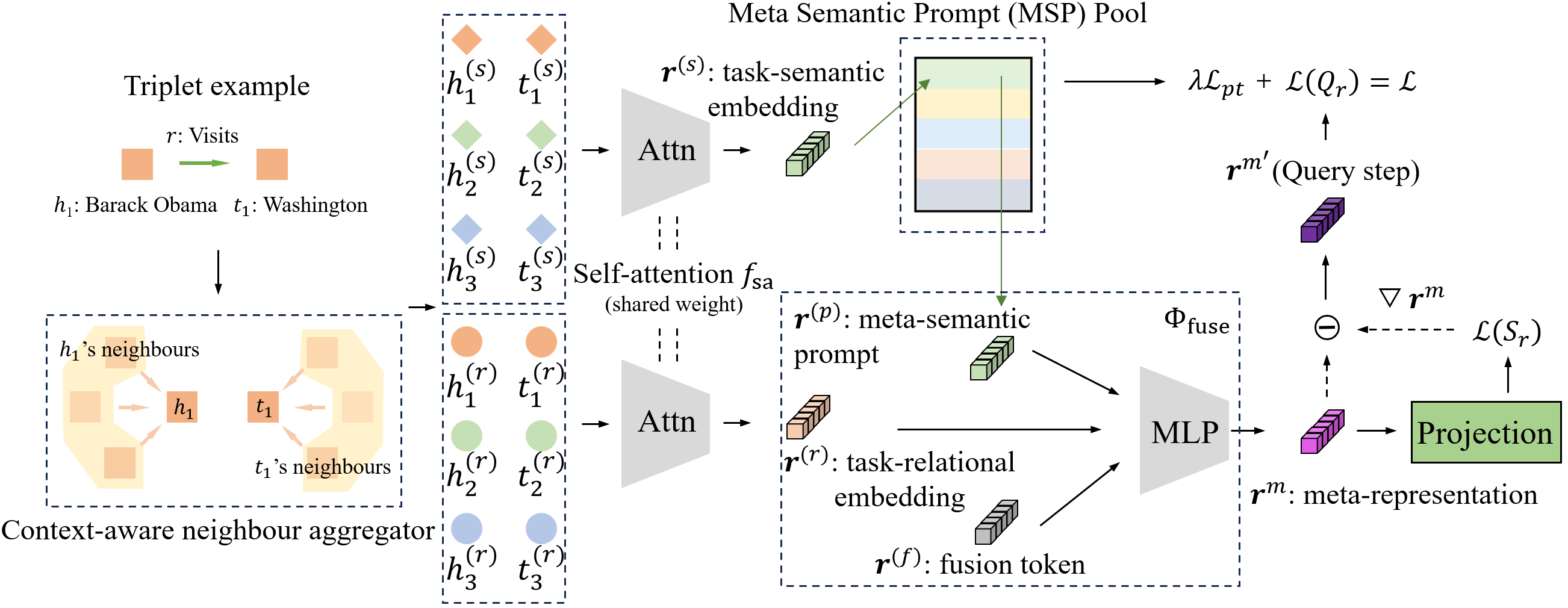}
    \vspace{-8pt}
    \caption{\small Illustration of  PromptMeta's meta-training process. Entities are initialized with pre-trained relational embeddings and enriched via context-aware neighbor aggregation to capture local relational information. The self-attention function $f_\textrm{sa}$ computes task-relational embedding $\mathbf{r}^{(r)}$, while task-semantic prompt $\mathbf{r}^{(p)}$ is retrieved from the MSP pool. These are fused through $\Phi_\textrm{fuse}$ with the fusion token $\mathbf{r}^{(f)}$ to generate the meta-representation $\mathbf{r}^m$. The support loss $\mathcal{L}(S_r)$ is computed to optimize model parameters, after which the updated meta-representation $\mathbf{r}^{m'}$ is adapted to the query set by optimizing the overall training loss $\mathcal{L}$.}
    \label{fig:flow}
\end{figure*}

\vspace{-1pt}
\subsection{Context-Aware Neighbor Aggregation}
As highlighted by prior studies~\cite{Gmatching, GANA}, local context information plays a vital role in learning entity embeddings in KGs. In light of this, we propose a context-aware neighbor encoder that aggregates local neighborhood of a given entity to enrich its embedding learning. 

For a given entity \( e \), its set of neighboring relation-entity tuples is defined as:  
$\mathcal{N}_e = \{(r, t) \mid (e, r, t) \in \mathcal{G}\} \cup \{(h, r) \mid (h, r, e) \in \mathcal{G}\}$. Our neighbor encoder enriches the embedding of entity \( e \) by leveraging relational information from its local neighborhood \( \mathcal{N}_e \). To explicitly capture relational directionality, we maintain separate embeddings for each relation \( r \) and its inverse \( r^{-1} \). If \( e \) is the head entity in \( (e, r, t) \), we concatenate the embeddings of \( r \) and \( t \). If \( e \) is the tail entity in \( (h, r, e) \), we concatenate the embeddings of \( r^{-1} \) and \( h \). 


For each neighboring tuple \( (r_i, e_i) \in \mathcal{N}_e \) (where both head and tail entities are uniformly represented as \( e_i \) for simplicity) is encoded as the concatenation of the relational embedding of relation \( r_i \) and entity \( e_i \), denoted as:  
$
\mathbf{n}^{(r)}_{i} = [\mathbf{r}^{(r)}_{i}; \mathbf{e}^{(r)}_{i}].
$
The relational embedding of the target entity \( e \), denoted as \( \mathbf{e}^{(r)} \), is updated by aggregating information from its local neighborhood:\vspace{-2pt}
\begin{align}
\mathbf{e}^{(r)'} = f_\textrm{n}\left(\sum_{i=1}^{\mid\mathcal{N}_e\mid} a_i \mathbf{n}^{(r)}_{i}\right) + \mathbf{e}^{(r)},
\label{eq:aggr}
\end{align}
where \( \mathbf{e}^{(r)} \in \mathbb{R}^{1 \times d} \) is the relational embedding of \( e \), \( a_i \) is the attention weight measuring the importance of the tuple embedding \( \mathbf{n}^{(r)}_{i} \) based on its relevance to \( e \), and \( f_\textrm{n} \) is a two-layer MLP. Neighboring tuples with higher attention weights contribute more significantly to the embedding of \( e \), thus enhancing the learning of local relational information.  

The attention weight \( a_i \) of the $i$-th neighboring tuple \( (r_i, e_i) \in \mathcal{N}_e\) is computed as follows:  
\begin{align}
\alpha_i &= g_\textrm{n}(\mathbf{e}^{(r)}\mathbf{W}_q, \mathbf{n}^{(r)}_{i}\mathbf{W}_k),\\
a_i &= \frac{\exp(\alpha_i)}{\sum_{j=1}^{\mid\mathcal{N}_e\mid} \exp(\alpha_j)},
\end{align}
where \( \mathbf{n}^{(r)}_{i} \in \mathbb{R}^{1\times 2d} \) is the relational embedding of the \( i \)-th tuple, and \( \mathbf{W}_q \in \mathbb{R}^{d\times d_k} \), \( \mathbf{W}_k \in \mathbb{R}^{2d\times d_k} \) are projection matrices. The function \( g_\textrm{n} \) performs cross-attention using the target entity embedding \( \mathbf{e}^{(r)} \) as the query on the neighboring tuples, implemented as a single-layer MLP with LeakyReLU activation. The raw attention scores \( \alpha_i \) are normalized via a softmax function to yield the final attention weights for local information aggregation.

\subsection{Learnable Meta-Semantic Prompt Pool}
KGs offer structured representations of real-world knowledge, inherently containing rich semantic information. However, most current FSRL works focus solely on relational structures, overlooking the semantic contexts embedded within entities and relations. This limitation is particularly pronounced in few-shot settings, where data scarcity hinders model generalization~\citep{challenging}. 
Moreover, current works lack mechanisms to explicitly learn and transfer semantic knowledge from seen to new relations. To bridge this gap, we propose a Meta-Semantic Prompt (MSP) pool, which serves as a repository of high-level semantic patterns learned across tasks. The MSP pool is randomly initialized and dynamically consolidated during meta-training to learn meta-semantic knowledge, enabling effective knowledge transfer.  

For a given few-shot task, we first construct a task-semantic embedding by aggregating semantic information from the support set. This embedding is then used to retrieve the most relevant meta-semantic prompt from the MSP pool, enriching the representation of the target relation and enabling rapid adaptation to unseen relations.  

\paragraph{Task-Semantic Embedding Aggregation.}  
Given a $K$-shot task $\mathcal{T}_{r}$ of relation $r$, we aggregate semantic information from the support set to construct a task-semantic embedding for $r$, denoted as $\mathbf{r}^{(s)}$. Specifically, we employ a self-attention mechanism to capture pairwise interactions among support triplets, allowing more informative triplets to contribute more significantly to $\mathbf{r}^{(s)}$. Formally, the task-semantic embedding $\mathbf{r}^{(s)}$ is computed as:  
\begin{align}
\mathbf{r}^{(s)} &= f_\textrm{sa}\left(\left\{[\mathbf{h}^{(s)}_{i};\mathbf{t}^{(s)}_{i}]\right\}_{i=1}^{K}\right),
\label{eq:fsa}
\end{align}  
where $[\mathbf{h}^{(s)}_{i};\mathbf{t}^{(s)}_{i}]$ denotes the concatenated semantic embeddings of the head and tail entities. The self-attention function $f_\textrm{sa}$ aggregates these embeddings by assigning greater attention weights to more representative triplets. 

\paragraph{Meta-Semantic Prompt Retrieval.}  
After aggregating the task-semantic embedding, we retrieve relevant meta-semantic knowledge from the MSP pool to provide semantic interpretation of relation $r$. Formally, let $\mathbf{P} \in \mathbb{R}^{M \times D_s}$ denote the MSP pool with $M$ components, where each component $\mathbf{p}_j \in \mathbb{R}^{1 \times D_s}$ represents a meta-semantic prompt vector of dimension $D_s$. The MSP pool encapsulates high-level meta-semantic patterns shared across tasks, enhancing generalization for FSRL.  


Given a task-semantic embedding $\mathbf{r}^{(s)}$, the corresponding meta-semantic prompt $\mathbf{r}^{(p)}$ is retrieved from the MSP pool based on the  criterion:
\begin{align}
    \mathbf{r}^{(p)} &= \argmax_{j \in \{1, 2, \ldots, M\}} \eta(\mathbf{r}^{(s)}, \mathbf{p}_j),
\end{align}    
where $\eta$ is a scoring function that measures cosine similarity between $\mathbf{r}^{(s)}$ and each pool component $\mathbf{p}_j$. The meta-semantic prompt with the highest similarity is selected to provide semantic cues relevant to the current task.

The MSP pool is dynamically updated after each meta-training task, retaining high-level semantics shared across tasks to facilitate effective knowledge transfer and generalization to unseen relations.  

\paragraph{Pool Tuning.}  
To optimize the learnable knowledge maintained in the MSP pool, we introduce a pool tuning objective based on the intuition that support triplets associated with the same relation should retrieve similar meta-semantic prompts, whereas triplets from different relations should retrieve more dissimilar ones. To this end, we adopt a contrastive learning-based approach to regularize the prompt retrieval process. For a given relation $r$, each support triplet and its retrieved meta-semantic prompt $\mathbf{r}^{(p)}$ form a positive pair, encouraging their embeddings to be close in the semantic space. In contrast, meta-semantic prompts retrieved for distinct relations are treated as negative pairs, promoting greater separation within the MSP pool.

This approach coherently consolidates meta-semantic prompts within the semantic space, allowing the model to leverage them as transferable knowledge for enhanced generalization and adaptation to unseen FSRL tasks.  

Formally, the pool tuning loss is defined as:
{\small
\begin{align}
    \mathcal{L}_\mathrm{pt} = \frac{1}{K}\sum_{i=1}^{K}-\log\frac{\exp\left(\eta(\mathbf{r}^{(p)}, [\mathbf{h}^{(s)}_{i};\mathbf{t}^{(s)}_{i}])/\tau\right)}{\sum\limits^{N}_{n=1}\exp\left(\eta(\mathbf{r}^{(p)}, {\mathbf{r}^{-}}^{(p)}_n)/\tau\right)},
    \label{eq:infonce}
\end{align}
}
where $N$ denotes the number of meta-semantic prompts retrieved for a different relation $r^-$ within the same batch.  $\tau$ is a temperature hyperparameter.  

\subsection{Relational and Meta-Semantic Information Fusion}
For a given task $\mathcal{T}_r$, we construct a joint meta-representation for relation $r$ by fusing two sources of knowledge: the task-relational embedding $\mathbf{r}^{(r)}$, and the meta-semantic knowledge encapsulated by $\mathbf{r}^{(p)}$. The task-relational embedding $\mathbf{r}^{(r)}$ is derived from the relational head and tail entity embeddings, which are first enriched via the context-aware neighbor aggregator and then aggregated via a self-attention function $f_\textrm{sa}$:
\begin{align}
    \mathbf{r}^{(r)} = f_\textrm{sa}\left(\left\{[\mathbf{h}^{(r)}_{i};\mathbf{t}^{(r)}_{i}]\right\}_{i=1}^{K}\right),
\end{align}
where $f_\textrm{sa}$ share parameters as in Eq.~\ref{eq:fsa}.

To adapt the task-relational embedding $\mathbf{r}^{(r)}$ with the meta-semantic knowledge from $\mathbf{r}^{(p)}$, we introduce a learnable fusion token $\mathbf{r}^{(f)}$, which serves as a flexible intermediary, dynamically adapting to the current task. The adapted meta-representation $\mathbf{r}^m$ for task $\mathcal{T}_r$ is then obtained as follows:
\begin{align}
\label{eq:mr_r}
\mathbf{r}^m = \Phi_\textrm{fuse}(\mathbf{r}^{(r)}, \mathbf{r}^{(p)}, \mathbf{r}^{(f)}),
\end{align}
where $\Phi_\textrm{fuse}$ is a fusion function implemented as a two-layer MLP. 
This flexible fusion scheme allows the model to dynamically adapt to task-relevant information, thereby ensuring a cohesive and well-integrated meta-representation. 

\subsection{Model Optimization}
After obtaining the meta-representation $\mathbf{r}^m$ from the support set, we adapt it to the query set using the optimization paradigm inspired by TransD~\citep{TransD}. For a target relation $r$ and each triplet $(h_i,r, t_i) \in \mathcal{S}_r$, we first obtain the head and tail entity embeddings, \( \mathbf{h}_i \) and \( \mathbf{t}_i \), by combining their relational and semantic components as follows:  
$
\mathbf{h}_i = \mathbf{h}^{(r)}_{i}+\mathbf{h}^{(s)}_{i},  \mathbf{t}_i = \mathbf{t}^{(r)}_{i}+\mathbf{t}^{(s)}_{i}.
$  
These embeddings are then projected into a latent space, jointly determined by the corresponding entities and relation. Specifically, the projection is defined as:  
\begin{align}
\mathbf{h}'_i &= \mathbf{M}_{h} \mathbf{h}_i, \label{eq:hpi}\quad \mathbf{M}_{h} = \mathbf{r}_{p} \mathbf{h}_{pi}^{\top} + \mathbf{I} ,\nonumber\\
\mathbf{t}'_i &= \mathbf{M}_{t} \mathbf{t}_i,\quad \mathbf{M}_{t} = \mathbf{r}_{p} \mathbf{t}_{pi}^{\top} + \mathbf{I},\nonumber
\end{align}
where \( \mathbf{h}'_i \) and \( \mathbf{t}'_i \) are the projected embeddings of the head and tail entities. \( \mathbf{I} \) is an identity matrix. The projection vectors $\textbf{r}_{p}$ and $\textbf{h}_{pi}$ (or $\textbf{t}_{pi}$) jointly determine the projection matrices \(\mathbf{M}_{h}\) for head entity $h_i$ (or \(\mathbf{M}_{t}\) for tail entity $t_i$), which are randomly initialized and updated during optimization.


\paragraph{Optimization on Support Set.}  
To measure the plausibility of each support triplet, we define a scoring function for each triplet $(h_i,r, t_i) \in \mathcal{S}_r$:
\begin{align}
\mathrm{score}(h_i, t_i) = \| \mathbf{h}'_{i} + \mathbf{r}^m - \mathbf{t}'_{i} \|_2,
\end{align}
\label{eq:score-support}
where $\|\cdot\|_2$ denotes the $\ell_2$ norm. 

To learn meta-representations, we minimize a margin-based ranking loss over the support set:
\begin{equation}
\small
\mathcal{L}(\mathcal{S}_r) = \sum_{\mathclap{(h_i,r,t_i) \in \mathcal{S}_r}}\max\big\{0, \mathrm{score}(h_i, t_i) + \gamma - \mathrm{score}(h_i, t_i^-)\big\},
\label{eq: marginloss}
\end{equation}
where $\gamma$ is a margin hyperparameter. The negative triplet $(h_i,r, t_i^-)$ is sampled such that $(h_i, r, t_i^-) \notin \mathcal{G}$. The meta-representation $\mathbf{r}^m$ is then refined based on the gradient of the support loss $\mathcal{L}(\mathcal{S}_r)$:
\begin{align}
\mathbf{r}^{m'} \leftarrow \mathbf{r}^m - \beta \nabla_{\mathbf{r}^m}\mathcal{L}(\mathcal{S}_r),
\label{eq: meta-update}
\end{align}
where $\beta$ is the learning rate. 
The projection vectors $\textbf{h}_{pi}$, $\textbf{r}_{p}$, and $\textbf{t}_{pi}$ are also updated similarly.

\paragraph{Adaptation to Query Set.}  
Using the updated meta-representation $\mathbf{r}^{m'}$ and projection vector $\textbf{r}_{p}$, we project each query triplet $(h_j, r, t_j) \in \mathcal{Q}_r$ and assess its plausibility with the same scoring function:
\vspace{-0.2cm}
\begin{align}
\mathrm{score}(h_j, t_j) &= \|\textbf{h}'_{j} + \mathbf{r}^{m'} - \textbf{t}'_{j}\|_2.
\end{align}
Accordingly, the query loss is calculated as:
\begin{equation}
    \small
    \mathcal{L}(\mathcal{Q}_r) = \sum_{\mathclap{(h_j, r, t_j) \in \mathcal{Q}_r}}\max\big\{0, \mathrm{score}(h_j, t_j) + \gamma - \mathrm{score}(h_j, t_j^-)\big\},   \label{eq:query-loss}
\end{equation}
$(h_j, r, t_j^-)$ is a negative triplet generated similarly to $(h_i, r, t_i^-) \in \mathcal{S}_r$. 

The overall training objective combines the query loss with the pool tuning loss $\mathcal{L}_{\mathrm{pt}}$:
\begin{align}
    \mathcal{L} = \mathcal{L}(\mathcal{Q}_r) + \lambda\mathcal{L}_{\mathrm{pt}},
\label{eq:loss}
\end{align}
where $\lambda$ is a trade-off hyperparameter that balances meta-representation adaptation to query set and the pool tuning objective. The MSP pool is updated accordingly as this objective is optimized. Figure~\ref{fig:flow} illustrates the meta-training process of PromptMeta. The detailed complexity analysis of PromptMeta is provided in Appendix~\ref{app:complexity}.

\section{Experiments}
\subsection{Experimental Setup}

\paragraph{Datasets.}
We evaluate our method\footnote{Code and data are available at: \url{https://github.com/alexhw15/PromptMeta}} on two widely used KG benchmarks for FSRL, Nell-One and Wiki-One, following the standard experimental setup introduced by GMatching~\citep{Gmatching}. For consistency, we adopt the commonly used splits of 51/5/11 on Nell-One and 133/16/34 on Wiki-One as training/validation/test relations, respectively.
The statistics of both datasets are provided in Table~\ref{tab:datasets}. In our setup, relations associated with more than 50 but less than 500 triplets are selected as few-shot relations. A background graph is constructed by excluding few-shot relations from training, validation, and test sets to provide the neighborhood context. For each few-shot relation, we use the candidate entity sets provided by GMatching for evaluation on both datasets. 

\begin{table}[ht]
\centering
\small
\caption{Statistics of KG benchmark datasets.}
\vspace{-0.2cm}
\label{tab:datasets}
\renewcommand\tabcolsep{4.5pt}
\begin{tabular}{c|cccc}
\toprule
Dataset & \#\,Relations & \#\,Entities & \#\,Triplets & \#\,Tasks\\
\midrule
Nell-One & 358 & 68,545    & 181,109   & 67\\
Wiki-One & 822 & 4,838,244 & 5,859,240 & 183\\
\bottomrule
\end{tabular}
\vspace{-0.2cm}
\end{table}

\paragraph{Metrics.}
For evaluation, we adopt two widely used metrics: MRR (mean reciprocal rank) and Hits@$k$ (with $k = 1, 5, 10$). MRR measures the mean reciprocal rank of the correct tail entities, and Hits@$k$ indicates the ratio of the correct tail entities that rank among the top $k$. We compare our method against baseline methods under the most common 1-shot and 5-shot settings. Unless otherwise stated, we report the results of our method using BERT-pretrained semantic embeddings.

\paragraph{Settings of Our Method.}
We initialize relational embeddings using the entity and relation embeddings pretrained by TransE~\citep{TransE} on both datasets, as released by GMatching~\citep{Gmatching}. Following prior works, we set the embedding dimension to $100$ for Nell-One and $50$ for Wiki-One. To prevent overfitting, we apply drop path with a drop rate of $0.2$. The maximum number of neighbors for a given entity is capped at 50, as in prior works. For all results except for the hyperparameter sensitivity test on $\lambda$ in Eq.~\ref{eq:loss}, $\lambda$ is set to $0.05$.
The margin hyperparameter $\gamma$ in Eq.~\ref{eq: marginloss} is set to $1$. We train the model using mini-batch gradient descent with a batch size of $1,024$ on both datasets. The Adam optimizer is used with a learning rate of $0.001$. Our method is evaluated on the validation set every $1,000$ steps, and the best model within $80,000$ steps is chosen based on MRR for testing. 

\paragraph{Pretrained Semantic Entity Embeddings.}
To align with the relational entity embeddings for initialization as provided by GMatching~\citep{Gmatching}, we construct a corresponding set of pre-trained semantic entity embeddings derived from textual information on Nell-One and Wiki-One. Specifically, we process and generate text descriptions for each entity on Nell-One based on the original entity names. For Wiki-One, we enrich text descriptions for entities following~\cite{challenging}. Subsequently, we derive an initialization vector for each entity on both datasets using pretrained weights from GloVe~\citep{glove} and BERT-base-uncased~\citep{bert}, followed by Smooth Inverse Frequency embeddings~\citep{SIF} to generate compact representations. To maintain consistency with relational embeddings used in prior FSRL works, the dimension of pretrained semantic embeddings is set to 100 for Nell-One and 50 for Wiki-One.

\begin{table*}[t]
\centering
\tabcolsep 1pt
\vspace{-0.2cm}
\caption{Comparison against state-of-the-art methods on Nell-One and Wiki-One. MetaR-I and MetaR-P indicate the In-train and Pre-train of MetaR~\citep{MetaR}, respectively. The best performance is highlighted in bold.}
\vspace{-0.1cm}
\small
\setlength{\tabcolsep}{2.7pt}
\stackunder{\resizebox{\linewidth}{!}{
\begin{tabular}{c|cc|cc|cc|cc|cc|cc|cc|cc}
    \toprule
    & \multicolumn{8}{c|}{\textbf{Nell-One}} & \multicolumn{8}{c}{\textbf{Wiki-One}} \\
    \midrule
    \multirow{2}{*}{\textbf{Methods}} & \multicolumn{2}{c}{MRR} & \multicolumn{2}{c}{Hits@1} & \multicolumn{2}{c}{Hits@5} & \multicolumn{2}{c|}{Hits@10} & \multicolumn{2}{c}{MRR} & \multicolumn{2}{c}{Hits@1} & \multicolumn{2}{c}{Hits@5} & \multicolumn{2}{c}{Hits@10} \\
     & 1-shot & 5-shot & 1-shot & 5-shot & 1-shot & 5-shot & 1-shot & 5-shot & 1-shot & 5-shot & 1-shot & 5-shot & 1-shot & 5-shot & 1-shot & 5-shot \\
    \midrule
    TransE & 0.105 & 0.168 & 0.041 & 0.082 & 0.111 & 0.186 & 0.226 & 0.345 & 0.036 & 0.052 & 0.011 & 0.042 & 0.024 & 0.057 & 0.059 & 0.090 \\
    TransH & 0.168 & 0.279 & 0.127 & 0.162 & 0.160 & 0.317 & 0.233 & 0.434 & 0.068 & 0.095 & 0.027 & 0.047 & 0.060 & 0.092 & 0.133 & 0.177 \\
    DistMult & 0.165 & 0.214 & 0.106 & 0.140 & 0.174 & 0.246 & 0.285 & 0.319 & 0.046 & 0.077 & 0.014 & 0.035 & 0.034 & 0.078 & 0.087 & 0.134 \\
    ComplEx & 0.179 & 0.239 & 0.112 & 0.176 & 0.212 & 0.253 & 0.299 & 0.364 & 0.055 & 0.070 & 0.021 & 0.030 & 0.044 & 0.063 & 0.100 & 0.124 \\
    KG-BERT & 0.191 & 0.238 & 0.122 & 0.172 & 0.224 & 0.257 & 0.303 & 0.388 & 0.062 & 0.088 & 0.028 & 0.040 & 0.046 & 0.071 & 0.125 & 0.159 \\
    \midrule
    GMatching & 0.185 & 0.201 & 0.119 & 0.143 & 0.260 & 0.264 & 0.313 & 0.311 & 0.200 & - & 0.120 & - & 0.272 & - & 0.336 & - \\ 
    MetaR-I & 0.250 & 0.261 & 0.170 & 0.168 & 0.336 & 0.350 & 0.401 & 0.437 & 0.193 & 0.221 & 0.152 & 0.178 & 0.233 & 0.264 & 0.280 & 0.302 \\
    MetaR-P & 0.164 & 0.209 & 0.093 & 0.141 & 0.238 & 0.280 & 0.331 & 0.355 & 0.314 & 0.323 & 0.266 & 0.270 & 0.375 & 0.385 & 0.404 & 0.418 \\
    MetaP$\dagger$ & 0.232 & - & 0.179 & - & 0.281 & - & 0.330 & - &
    - & - & - & - & - & -  & - & - \\
    FSRL & - & 0.184 & - & 0.136 & - & 0.234 & - & 0.272 & - & 0.158 & - & 0.097 & - & 0.206 & - & 0.287 \\
    FAAN & - & 0.279 & - & 0.200 & - & 0.364 & - & 0.428 & - & 0.341 & - & 0.281 & - & 0.395 & - & 0.436 \\
    GANA & 0.236 & 0.245 & 0.173 & 0.166 & 0.293 & 0.334 & 0.347 & 0.390 & 0.260 & 0.261 & 0.221 & 0.317 & 0.307 & 0.333 & 0.334 & 0.384 \\
    HiRe & 0.288 & 0.306 & 0.184 & 0.207 & 0.403 & 0.439 & 0.472 & 0.520 & 0.322 & 0.371 & 0.271 & 0.319 & 0.383 & 0.419 & 0.433 & 0.469 \\
    RelAdapter$\dagger$ & - & - & - & - & - & -  & - & - & 0.247 & 0.305 & 0.209 & 0.245 & 0.281 & 0.365 & 0.307 & 0.415 \\
    \midrule
    PromptMeta & \textbf{0.293} & \textbf{0.338} & \textbf{0.194} & \textbf{0.239} & \textbf{0.409} & \textbf{0.445} & \textbf{0.475} & \textbf{0.526} & \textbf{0.339} & \textbf{0.392} & \textbf{0.284} & \textbf{0.335} & \textbf{0.392} & \textbf{0.433} & \textbf{0.436} & \textbf{0.480} \\
    \bottomrule
\end{tabular}
}}
{\parbox{6in}{
\small $\dagger$: The results of MetaP and RelAdapter are not fully available due to unreleased pretrained information. }
}
\label{tab:sota}
\end{table*}

\subsection{Baseline Methods}
We compare our method with two groups of state-of-the-art (SOTA) methods: (1) \textbf{KG embedding methods} that focus on modeling relational structures, with some also incorporating semantics in KGs. This group includes four classic methods, TransE~\citep{TransE}, TransH~\citep{TransH}, DistMult~\citep{DistMult}, and ComplEx~\citep{ComplEx}, as well as KG-BERT~\citep{KG-bert}, which jointly models relational and semantic information.
We use OpenKE~\citep{openke} to reproduce the results of these models with hyperparameters reported in the original papers. (2) \textbf{Few-shot relational learning methods}: For fair comparison, we compare against the SOTA FSRL baselines that follow the same training and evaluation protocol, including GMatching~\citep{Gmatching}, MetaR~\citep{MetaR}, FAAN~\citep{FAAN}, FSRL~\citep{FSRL}, GANA~\citep{GANA},  HiRe~\citep{HiRe}, and RelAdaper~\cite{RelAdapter}. All results are obtained after the models are trained using triplets from the background graph and meta-training tasks and evaluated on meta-testing tasks. All models are implemented by PyTorch and trained on a V100 GPU. Refer to Appendix~\ref{app:repro} for reproducibility details.

\begin{table*}[t]
\centering
\setlength{\tabcolsep}{1.8pt}
\caption{Ablation study of PromptMeta under 3-shot and 5-shot settings on Nell-One and Wiki-One. H@1/5/10 is short for Hits@1/5/10. The best performance is marked in grey.}
\small
\vspace{-0.2cm}
\scalebox{0.95}{
\begin{tabular}{c|cccc|cccc|cccc|cccc}
    \toprule
    & \multicolumn{8}{c|}{\textbf{Nell-One}} & \multicolumn{8}{c}{\textbf{Wiki-One}} \\
    \midrule
    \multirow{2}{*}{\textbf{Ablation on}} & \multicolumn{4}{c|}{3-shot} & \multicolumn{4}{c|}{5-shot} & \multicolumn{4}{c|}{3-shot} & \multicolumn{4}{c}{5-shot} \\
     & MRR & H@1 & H@5 & H@10 & MRR & H@1 & H@5 & H@10 & MRR & H@1 & H@5 & H@10 & MRR & H@1 & H@5 & H@10 \\
    \midrule
    w/o N.A. & 0.309 & 0.207 & 0.426 & 0.493 & 0.316 & 0.221 & 0.429 & 0.510 & 0.359 & 0.312 & 0.399 & 0.448 & 0.371 & 0.298 & 0.416 & 0.459 \\
    w/o Semantic         & 0.281 & 0.185 & 0.380 & 0.461 & 0.294 & 0.205 & 0.383 & 0.471 & 0.342 & 0.287 & 0.394 & 0.436 & 0.349 & 0.289 & 0.404 & 0.448 \\
    w/o MSP-Task-Sem. & 0.315 & 0.209 & 0.424 & 0.502 & 0.320 & 0.225 & 0.429 & 0.515 & 0.358 & 0.312 & 0.409 & 0.456 & 0.382 & 0.325 & 0.421 & 0.466 \\
    w/o Fusion Token     & 0.304 & 0.202 & 0.414 & 0.485 & 0.314 & 0.212 & 0.421 & 0.500 & 0.353 & 0.299 & 0.397 & 0.448 & 0.363 & 0.308 & 0.413 & 0.459 \\
    w/o Pool Tuning       & 0.311 & 0.211 & 0.419 & 0.494 & 0.315 & 0.219 & 0.418 & 0.508 & 0.360 & 0.314 & 0.405 & 0.452 & 0.379 & 0.320 & 0.418 & 0.462 \\
    \midrule
    PromptMeta (GloVe)   & 0.318 & 0.219 & 0.428 & 0.510 & 0.327 & 0.233 & 0.442 & 0.522 & 0.368 & 0.320 & 0.415 & 0.461 & 0.386 & 0.327 & 0.426 & 0.472 \\
    \rowcolor{gray!18}
    PromptMeta (BERT)    & 0.323 & 0.221 & 0.435 & 0.516 & 0.338 & 0.239 & 0.445 & 0.526 & 0.372 & 0.326 & 0.413 & 0.459 & 0.392 & 0.335 & 0.433 & 0.480 \\
    \bottomrule
\end{tabular}}
\vspace{-0.1cm}
\label{tab:ablation}
\end{table*}

\begin{table*}[t]
\centering
{\begin{subfigure}{0.32\linewidth}
    \hfill
    \setlength{\tabcolsep}{4pt}
    \footnotesize
    \begin{tabular}{c|ccc}
    \multirow{2}{*}{$\lambda$} & \multicolumn{3}{c}{\textbf{5-shot}} \\
         & MRR & H@1 & H@10 \\ 
        \hline
        0 & 0.317 & 0.220 & 0.509 \\
        0.01 & 0.332 & 0.234 & 0.518 \\        
        \cellcolor{lightgray}{0.05} & \cellcolor{lightgray}{0.338} & \cellcolor{lightgray}{0.239} & \cellcolor{lightgray}{0.526} \\
        0.10 & 0.330 & 0.229 & 0.522 \\
    \end{tabular}
    \hfill
    \vspace{-0.15cm}
    \caption{\small{Pool tuning weight.}}
    \label{tab:contras}
\end{subfigure}
}
\hspace{0.01em}
{\begin{subfigure}{0.32\linewidth}
    \hfill
    \setlength{\tabcolsep}{4pt}
    \footnotesize
    \begin{tabular}{c|ccc}
    \multirow{2}{*}{Samples} & \multicolumn{3}{c}{\textbf{5-shot}} \\
         & MRR & H@1 & H@10 \\ 
        \hline
        0 & 0.317 & 0.220 & 0.509 \\
        256 & 0.332 & 0.229 & 0.521 \\
        512 & 0.337 & 0.236 & 0.523 \\
        \cellcolor{lightgray}1,024 & \cellcolor{lightgray}0.338 & \cellcolor{lightgray}0.239 & \cellcolor{lightgray}0.526 \\
    \end{tabular}
    \hfill
    \vspace{-0.1cm}
    \caption{\small{Number of negative samples.}}
    \label{tab:negative}
\end{subfigure}
}
\hspace{.01em}
{\begin{subfigure}{0.32\linewidth}
    \hfill
    \setlength{\tabcolsep}{4pt}
    \footnotesize
    \begin{tabular}{c|ccc}
    \multirow{2}{*}{Pool size} & \multicolumn{3}{c}{\textbf{5-shot}} \\
         & MRR & H@1 & H@10 \\ 
        \hline
        0 & 0.298 & 0.208 & 0.479 \\
        32 & 0.323 & 0.230 & 0.515 \\
        \cellcolor{lightgray}64 & \cellcolor{lightgray}0.338 & \cellcolor{lightgray}0.239 & \cellcolor{lightgray}0.526 \\
        128 & 0.319 & 0.220 & 0.514 \\
    \end{tabular}
    \hfill
    \caption{\small{Size of the MSP pool.}}
    \label{tab:msp}
\end{subfigure}
}
\caption{Hyperparameter analysis on a) prompt tuning weight; b) the number of negative samples in contrastive loss; c) the size of MSP pool. $5$-shot setting on Nell-One are reported. Default settings are marked in gray. \label{tab:ablation_hyper}}
\vspace{-0.1cm}
\end{table*}

\subsection{Comparison with State-of-the-Art}
Table~\ref{tab:sota} compares our method with baseline methods on Nell-One and Wiki-One under $1$-shot and $5$-shot settings. Traditional KG embedding methods, \eg, TransE, TransH, and ComplEx, are designed for scenarios with abundant training data and thus perform significantly worse in few-shot settings. PromptMeta consistently outperforms SOTA FSRL methods like GANA and HiRe across both datasets and settings. On Nell-One, our method surpasses HiRe, the second-best approach, in terms of MRR and Hits@1 by 1.7\% and 5.4\% in the $1$-shot setting, and by 10.5\% and 15.5\% in the $5$-shot setting. As the number of support triplets increases from $1$-shot to $5$-shot, our method achieves even greater performance gains over baselines, largely attributed to the meta-semantic prompt pool that effectively captures and leverages shared meta-semantic information across different few-shot tasks.

\subsection{Ablation Studies}
To assess the contribution of each component in PromptMeta, we conduct a thorough ablation study by comparing it with four variants: (1) \textbf{w/o N.A.} that removes the first term of neighbor aggregation in Eq.~\ref{eq:aggr}; (2) \textbf{w/o Semantic} that ablates both $\mathbf{r}^{(p)}$ and $\mathbf{r}^{(f)}$ from Eq.~\ref{eq:mr_r}, indicating only task-relational embeddings are used to generate the meta-representation; (3) \textbf{w/o MSP-Task-Sem.}~that replaces the selected meta-semantic prompt $\mathbf{r}^{(p)}$ in Eq.~\ref{eq:mr_r} with task-semantic embedding $\mathbf{r}^{(s)}$ from Eq.~\ref{eq:fsa}; and (4) \textbf{w/o Fusion Token} that removes the fusion token $\mathbf{r}^{(f)}$ from Eq~\ref{eq:mr_r}; (4) \textbf{w/o Pool-tuning} is the variant without pool tuning. The results on Nell-One and Wiki-One under 3-shot and 5-shot settings are reported in the top panel of Table~\ref{tab:ablation}. We can find that our proposed mechanisms for integrating semantic information all contribute significantly to the overall results. Notably, the removal of the MSP pool and/or fusion token leads to a profound performance drop under both settings. This confirms the crucial role of the MSP pool in learning high-level, generalizable semantic knowledge and the utility of fusion token in integrating task-relational embedding with meta-semantic prompts for effective few-shot adaptation.





\subsection{The Effect of Semantic Embeddings}
We further compare the performance of our method using different semantic embeddings. As shown in the bottom panel of Table~\ref{tab:ablation}, 
PromptMeta (BERT) consistently outperforms its GloVe-based counterpart on both datasets. This performance gain stems from BERT's ability to capture contextualized word meanings, offering richer semantic representations of entities compared to GloVe's reliance on word co-occurrence statistics. These findings highlight the benefits of incorporating richer semantics, shedding light on future directions in exploring more advanced language models for FSRL in KGs. 

\subsection{Hyperparameter Analysis}
We report a sensitivity analysis on Nell-One in Table~\ref{tab:ablation_hyper} on three hyperparameters: the trade-off hyperparameter $\lambda$ in Eq.~\ref{eq:loss},
the number of negative samples $N$ in Eq.~\ref{eq:infonce},  and the MSP pool size $M$.

To account for different scales of the margin loss and the pool tuning loss, we study the impact of $\lambda$ in the range $[0, 0.1]$. As shown in Table~\ref{tab:contras}, PromptMeta achieves the best performance when $\lambda = 0.05$. The performance of PromptMeta consistently improves when $\lambda > 0$, validating the efficacy of our contrastive learning based pool tuning.

For the number of negative samples $N$, we vary its value from $0$ to $1,024$ in Table~\ref{tab:negative}. PromptMeta performs the best when $N=1,024$. The performance improvements with an increase in $N$ suggest that a larger negative sample set helps enhance the differentiation of meta-semantics.


Similarly, we examine the effect of the MSP pool size $M$ by varying it from $0$, $32$, $64$, to $128$. As shown in Table~\ref{tab:msp}, the best performance on Nell-One is observed when $M=64$. However, increasing $M$ to $128$ leads to a significant drop, likely due to increased overfitting. 
A moderate pool size effectively captures the shared semantic patterns essential for knowledge transfer, striking a good balance between capacity and generalization.

\subsection{Case Study}
\begin{figure}
    \centering   \includegraphics[width=\linewidth]{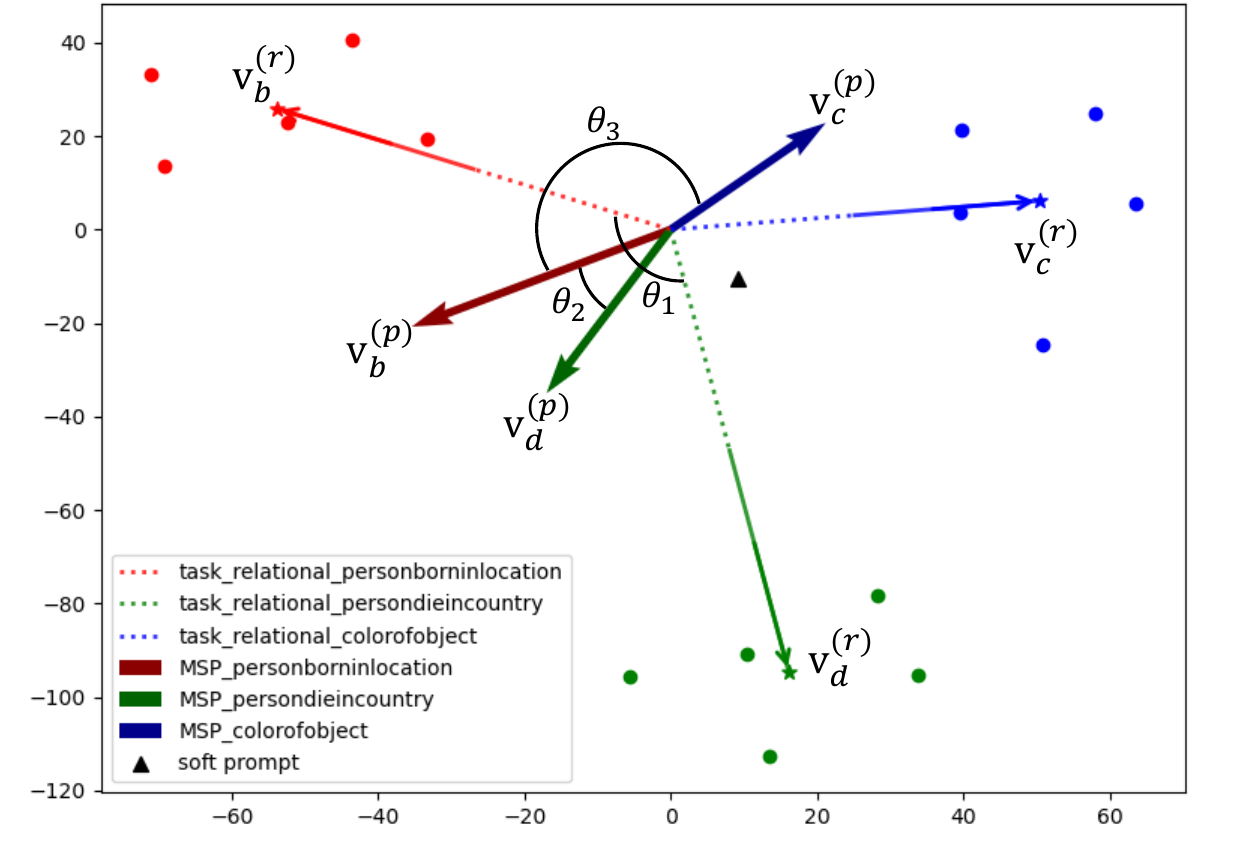}
    \caption{t-SNE visualization and case study.}
    \label{fig:tsne}
\end{figure}

Lastly, we present a case study to visually illustrate the efficacy of PromptMeta in augmenting relational embeddings with meta-semantic information for FSRL.  Figure~\ref{fig:tsne} shows the t-SNE visualization of three relations from Nell-One: \texttt{color\_of\_object}, \texttt{person\_born\_in\_location}, and \texttt{person\_die\_in\_country}, under the 5-shot setting. In the figure, dotted arrows indicate relational embeddings conventionally used in FSRL, whereas solid arrows denote meta-semantic prompt (MSP) vectors learned by PromptMeta. As expected, the MSP vectors for the semantically related relations \texttt{person\_born\_in\_location} and \texttt{person\_die\_in\_country} exhibit greater similarity, as evidenced by a smaller angle $\theta_2$ between them. This angle is notably smaller than $\theta_3$, the angle formed by the MSP vectors of the dissimilar pair \texttt{person\_die\_in\_country} and \texttt{color\_of\_object}. This observation affirms the ability of PromptMeta to learn more generalizable semantic patterns shared across similar relations. By incorporating the learned MSP vectors to augment relational embeddings, PromptMeta effectively reduces the angular distance ($\theta_1$) between the embeddings of semantically related relations, thereby improving generalization across few-shot relations. Please refer to Appendix~\ref{app:case-study} for a quantitative case study on the benefits of utilizing semantic information for FSRL.


\section{Conclusion}
This paper presents PromptMeta, a novel prompted meta-learning framework for FSRL. By integrating meta-semantic knowledge with relational information, PromptMeta offers advantageous knowledge transfer and model adaptation for predicting unseen relations in few-shot settings. Experiments on two KG benchmarks validate the superiority of PromptMeta over state-of-the-art FSRL methods. The ablation study and hyperparameter analysis further verify the significance of its key components. Moreover, case studies on real-world KG datasets highlight the benefits of incorporating meta-semantic knowledge to improve model generalization.

\section{Limitations}

The proposed PromptMeta framework demonstrates competitive performance by incorporating meta-semantic knowledge into FSRL, but several potential limitations warrant further discussion. First, the MSP pool is designed to learn high-level semantic patterns shared across tasks, its effectiveness depends on the presence of semantically related relations during meta-training. The model's ability to learn generalizable meta-semantics may be limited when trained on a set of highly diverse or unrelated relations. 
Second, PromptMeta's performance gains are influenced by the quality of pretrained semantic embeddings. While this work utilizes GloVe~\citep{glove} and BERT-base-uncased~\citep{bert}, future work will investigate more advanced pretrained language models, including large language models, to generate richer and more expressive semantic embeddings. Finally, the use of semantic information, though beneficial for performance, introduces additional memory cost. Exploring memory-efficient solutions will be important for extending the applicability of PromptMeta to larger-scale KGs under few-shot scenarios. 


\bibliography{fewshot,custom}

\newpage
\appendix
\noindent\textbf{\large{Appendices}}

\section{Complexity Analysis}
\label{app:complexity}

We analyze the computational complexity of our proposed PromptMeta for its key components. Table~\ref{tab:complexity} presents the complexity of PromptMeta in terms of the number of model parameters and the number of multiplication operations per task. Herein, $d$ denotes the dimension of entity embeddings (both relational and semantic), $n$ denotes the number of  neighbors per entity, and $K$ denotes the number of support triplets in each task. The overall complexity of PromptMeta grows quadratically with the number of support triplets $K$ in each task and the number of neighbors $n$ per entity. However, our proposed PromptMeta remains reasonably scalable in practice. Since few-shot tasks involve a very small number of support triplets $K$  (typically 1, 3 or 5) and a fixed number of neighbors $n$ (\ie, 50), the overall complexity of PromptMeta is comparable to other state-of-the-art methods.

\section{Reproducibility Details}
\label{app:repro}

We compare our method and all baselines under the same evaluation framework. The results for the baseline methods are either adopted from their respective papers, or reproduced using their officially released open-source implementations. For MetaR~\cite{MetaR} (both In-Train and Pre-Train)\footnote{MetaR: \url{https://github.com/AnselCmy/MetaR}}, FAAN~\cite{FAAN}\footnote{FAAN: \url{https://github.com/JiaweiSheng/FAAN}}, and HiRe~\cite{HiRe}\footnote{HiRe: \url{https://github.com/alexhw15/HiRe}}, we directly adopt the results reported in their respective papers. The results for GMatching~\cite{Gmatching}\footnote{GMatching:\url{https://github.com/xwhan/One-shot-Relational-Learning}} under 1-shot and 5-shot settings are taken from~\cite{MetaR}. Since FSRL~\cite{FSRL}\footnote{FSRL: \url{https://github.com/chuxuzhang/AAAI2020\_FSRL}} was originally evaluated under different conditions (with a smaller candidate set), we use the re-implemented results provided by FAAN to ensure consistency. For GANA~\cite{GANA}\footnote{GANA: \url{https://github.com/ngl567/GANA-FewShotKGC}}, we replicate its results by setting the number of neighbors to 
50, in line with the configurations of other methods. We also reproduce the results of KG-BERT~\cite{KG-bert}\footnote{KG-BERT: \url{https://github.com/yao8839836/kg-bert}} and RelAdapter~\cite{RelAdapter}\footnote{RelAdapter: \url{https://github.com/smufang/RelAdapter}} under the 1-shot and 5-shot settings on Wiki-One. However, as the pretrained contextual information required for Nell-One is not released by RelAdapter, we are unable to evaluate its performance on this dataset. All models are implemented in PyTorch and trained on a single Nvidia V100 GPU.

\section{A Quantitative Case Study on the Benefits of Semantic Information}
\label{app:case-study}

To quantitatively evaluate the benefits of exploiting meta-semantic information to improve FSRL, we conduct a case study on two semantically related relations from the Nell-One dataset: \texttt{Automobilemaker-Dealer-In-City} and \texttt{Automobilemaker-Dealers-In-Country}. The two relations were chosen because they are semantically close, making them an appropriate case for evaluating whether our model can leverage semantic cues to   distinguish and generalize between such similar relations. 

We compare our model, PromptMeta, against the best-performing baseline, HiRe, in a challenging 1-shot setting. PromptMeta was tested using entity embeddings pretrained with both GloVe and the more semantically rich BERT. The comparison results are reported in Table~\ref{tab:quantative}. 

We can observe that, for the relation \texttt{Automobilemaker-Dealer-In-City}, both methods produce comparable results, with PromptMeta showing a slight advantage. However, a more pronounced difference emerges when we examine the relation \texttt{Automobilemaker-Dealers-In-Country}. As compared with HiRe, PromptMeta boosts the MRR from 0.164 to 0.265. This substantial improvement affirms the ability of PromptMeta to effectively utilize the shared semantic similarity from the description of entities across the two relations.

Furthermore, it is evident that utilizing a stronger semantic embedding pretrained with BERT, as seen in PromptMeta, leads to a marked performance gain for the relation \texttt{Automobilemaker-Dealers-In-Country}. This suggests that PromptMeta benefits from its ability to capture and utilize latent semantic cues, which is critical to distinguish subtle differences between semantically related relations.

In summary, this case study demonstrates that explicitly modeling semantic information is crucial for improving FRSL, particularly for relations that share strong contextual similarities. 

\clearpage            
\thispagestyle{empty} 
\vspace*{0pt}       

    \noindent\makebox[\textwidth][c]{%
    \begin{minipage}{0.8\textwidth} 
    \centering
    \captionof{table}{Computational complexity of PromptMeta. N.A. is short for neighbor aggregation. $d$ denotes the dimension of entity embeddings. $n$ denotes the number of neighbors per entity, and $K$ denotes the number of support triplets in each task, where $K \ll n, d$. }
    \vspace{-0.2cm}
    \renewcommand\tabcolsep{15pt}
    \begin{tabular}{lcc}
        \toprule
        & \textbf{\# Parameters} & \textbf{\# Multiplication Operations} \\
        \midrule
        N.A. & $\mathcal{O}(nKd + n^2K)$ & $\mathcal{O}(nKd^2 + n^2Kd)$ \\
        MSP & $\mathcal{O}(Kd^2)$   & $\mathcal{O}(Kd^2 + K^2d)$ \\
        Fusion & $\mathcal{O}(Kd + K^2)$ & $\mathcal{O}(Kd^2 + K^2d)$ \\
        \midrule
        Total & $\mathcal{O}(nKd + Kd^2 + n^2K)$ & $\mathcal{O}(nKd^2 + n^2Kd)$ \\
        \bottomrule
    \end{tabular}  
    \label{tab:complexity}
    \end{minipage}
}

\bigskip
\bigskip
    
    \noindent\makebox[\textwidth][c]{%
    \begin{minipage}{\textwidth} 
    \centering
    \captionof{table}{A quantitative case study on the benefits of utilizing semantic information.}
    \vspace{-0.2cm}
    \setlength{\tabcolsep}{2.6pt}
    \begin{tabular}{c|cccc|cccc}
    \toprule
    \multirow{2}{*}{Model} & \multicolumn{4}{c|}{\texttt{Automobilemaker-Dealer-In-City}} & \multicolumn{4}{c}{\texttt{Automobilemaker-Dealers-In-Country}} \\
    \cmidrule(lr){2-5} \cmidrule(lr){6-9}
    & MRR & Hits@1 & Hits@5 & Hits@10 & MRR & Hits@1 & Hits@5 & Hits@10 \\
    \midrule
    HiRe               & 0.604 & 0.538 & 0.663 & 0.721 & 0.164 & 0.104 & 0.214 & 0.283 \\
    PromptMeta (GloVe) & 0.618 & 0.505 & 0.788 & 0.894 & 0.231 & 0.159 & 0.338 & 0.475 \\
    PromptMeta (BERT)  & 0.646 & 0.516 & 0.835 & 0.912 & 0.265 & 0.137 & 0.425 & 0.562 \\
    \bottomrule
    \end{tabular}
    \label{tab:quantative}
\end{minipage}
}

\end{document}